\documentclass{article}

\usepackage[preprint, nonatbib]{neurips_data_2024}

\usepackage[utf8]{inputenc} 
\usepackage[T1]{fontenc}    
\usepackage{hyperref}       
\usepackage{url}            
\usepackage{booktabs}       
\usepackage{amsfonts}       
\usepackage{nicefrac}       
\usepackage{microtype}      
\usepackage{xcolor}         

\usepackage{multirow}
\usepackage{graphicx}
\usepackage{subfigure}
\usepackage{pifont}
\usepackage{amsmath}
\usepackage{hyperref}

\def\ie{{\em i.e. }}
\def\eg{{\em e.g. }}
\def\etal{{\em et al. }}

\def\yes{\textcolor{green}{\ding{51}}}
\def\no{\textcolor{red}{\ding{55}}}

\title{\textit{RoboCAS}: A Benchmark for Robotic Manipulation in Complex Object Arrangement Scenarios} 

\author{
    Liming Zheng \\
    Meituan \\
    Beijing, China 100012 \\
    \texttt{zhengliming04@meituan.com} \\
    \And
    Feng Yan \\
    Meituan \\
    Beijing, China 100012 \\
    \texttt{yanfeng05@meituan.com} \\
    \And
    Fanfan Liu \\
    Meituan \\
    Beijing, China 100012 \\
    \texttt{liufanfan03@meituan.com} \\
    \And
    Chengjian Feng \\
    Meituan \\
    Shenzhen, China 518110 \\
    \texttt{fengchengjian@meituan.com} \\
    \And
    Zhuoliang Kang \\
    Meituan \\
    Beijing, China 100012 \\
    \texttt{kangzhuoliang@meituan.com} \\
    \And
    Lin Ma$^{\dag}$ \\
    Meituan \\
    Beijing, China 100012 \\
    \texttt{forest.linma@gmail.com}
}

\begin{document}

\maketitle

\begin{abstract}
  Foundation models hold significant potential for enabling robots to perform long-horizon general manipulation tasks. However, the simplicity of tasks and the uniformity of environments in existing benchmarks restrict their effective deployment in complex scenarios. To address this limitation, this paper introduces the \textit{RoboCAS} benchmark, the first benchmark specifically designed for complex object arrangement scenarios in robotic manipulation. This benchmark employs flexible and concise scripted policies to efficiently collect a diverse array of demonstrations, showcasing scattered, orderly, and stacked object arrangements within a highly realistic physical simulation environment. It includes complex processes such as target retrieval, obstacle clearance, and robot manipulation, testing agents' abilities to perform long-horizon planning for spatial reasoning and predicting chain reactions under ambiguous instructions. Extensive experiments on multiple baseline models reveal their limitations in managing complex object arrangement scenarios, underscoring the urgent need for intelligent agents capable of performing long-horizon operations in practical deployments and providing valuable insights for future research directions. Project website: \url{https://github.com/notFoundThisPerson/RoboCAS-v0}.

\end{abstract}
{\let\thefootnote\relax\footnotetext{$^{\dag}$ Corresponding authors.}}

\section{Introduction}

In the field of artificial intelligence (AI), embodied AI~\cite{salvato2021crossing, brohan2023can, rt-x} is increasingly becoming a focal point of research. Its core objective is to develop intelligent systems that can deeply understand their environments, make precise decisions, and perform complex physical operations. To achieve this high level of intelligence, researchers utilize advanced methods such as imitation learning ~\cite{3d_diffuser_actor, act3d, dp2023} and reinforcement learning ~\cite{salvato2021crossing, elguea2023review, mandlekar2021matters, apolinarska2021robotic}, which have proven effective in multiple experiments and applications. However, the successful implementation of these technologies heavily relies on a large amount of high-quality training data, which is often challenging to obtain.

\begin{figure}[t]
    \centering
    \includegraphics[width=0.9\linewidth, trim=110 130 110 130, clip]{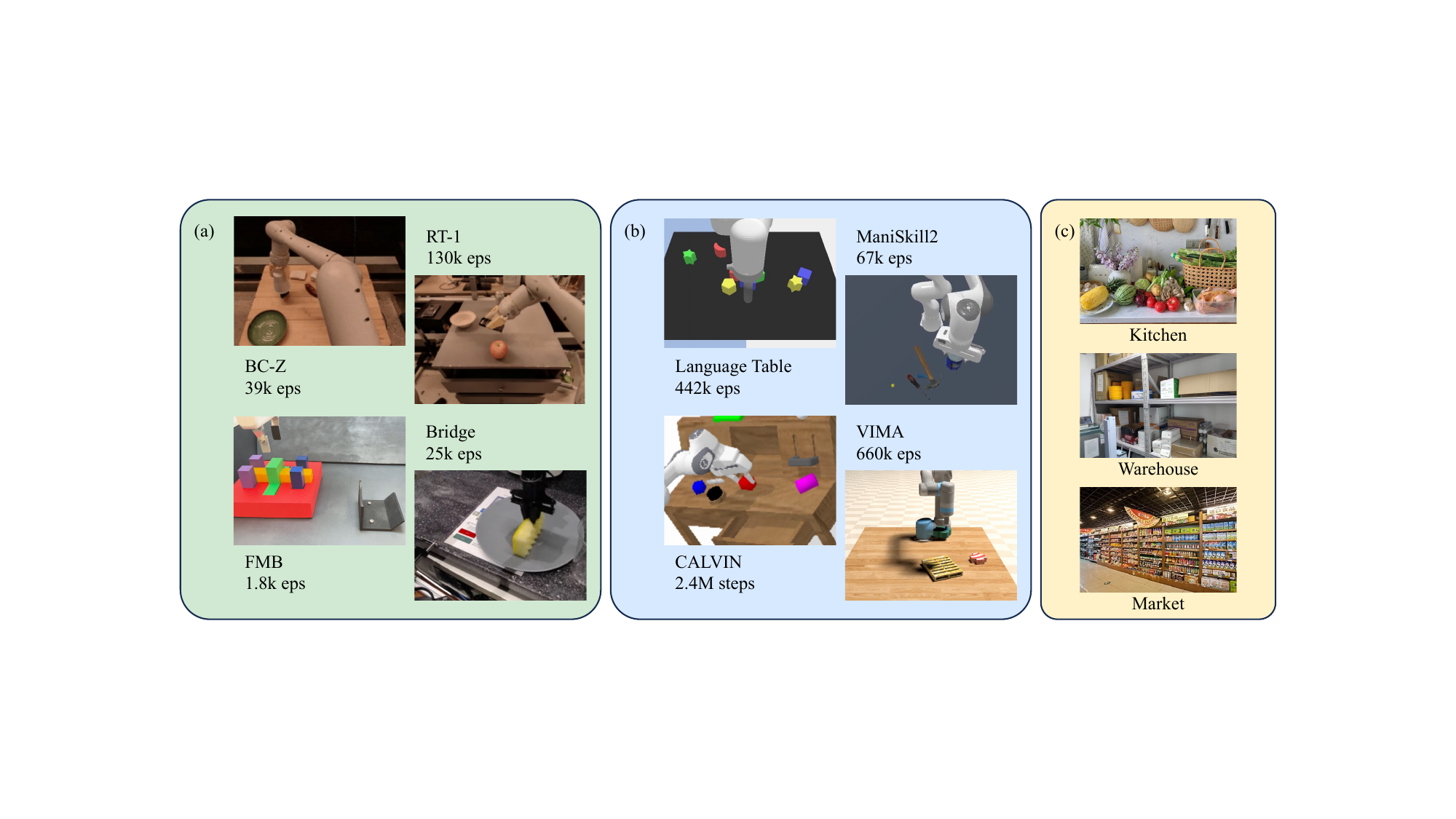}
    \caption{Visualization. (a) Real-world datasets: The amount of data collected on real robots is relatively small, and the scene layouts are quite simple. (b) Simulation datasets: Although there is a large volume of data, the environments are monotonous and the tasks are simple. (c) Real-world scenes: The complexity of object placement far exceeds that in (a) and (b). 
    }
    \label{fig:datasets}
\end{figure}

Currently, researchers primarily rely on two types of datasets to train and test these intelligent systems: real-world robot datasets and simulation datasets. Real-world robot datasets~\cite{rt-x, aloha, mobile-aloha, bc-z, luo2024fmb} provide real-world scenes and physical interaction data, which are extremely beneficial for model training. However, the production of these datasets involves high costs, including the procurement of expensive robot hardware, setting up complex environments, and extensive data collection and annotation efforts. Additionally, these datasets have long production cycles; for example, the RT-1 dataset~\cite{rt12022arxiv} collected only 130k expert demonstrations over 17 months. Due to these reasons, the existing benchmarks have limited data volumes and relatively simple tasks, as shown in Figure~\ref{fig:datasets}(a).

Meanwhile, simulation datasets offer a cost-effective alternative. Researchers use platforms such as Gazebo~\cite{koenig2004design} and IsaacGym~\cite{makoviychuk2021isaac} to simulate various scenarios and tasks through computer simulations, rapidly generating large volumes of data. However, current simulation data often lack the complexity and diversity of the real world, which frequently becomes a major barrier in the transition from simulation to reality (Sim2Real). As illustrated in Figure~\ref{fig:datasets}(b), existing datasets primarily focus on clean and tidy scenes, such as monotonous desktops and backgrounds, and involve tasks like grasping non-realistic objects, such as picking up blocks.

As illustrated Figure~\ref{fig:datasets}(c), everyday life commonly features objects that are orderly arranged and stacked, such as piles of vegetables or neatly arranged goods on shelves. These scenarios require agents to handle ambiguous language instructions, perform advanced spatial reasoning, and predict chain reactions. Based on this, we propose a new benchmark named "Robotic Manipulation in Complex Object Arrangement Scenarios" (\textit{RoboCAS}). This benchmark utilizes flexible and concise scripted policies to efficiently collect a wide array of demonstrations showcasing scattered, orderly, and stacked object placements in a highly realistic physical simulation environment. These demonstrations cover complex processes such as target retrieval, obstacle clearance, and robot manipulation. Notably, the objects in the environment, such as tables and wardrobes, as well as manipulable objects like cups and facial cleansers, are all sourced from scans of real objects. Additionally, the developed scripted policies automate the generation of scenes and demonstration trajectories, enabling the cost-effective collection of training and validation data for imitation learning.

This benchmark aims to comprehensively evaluate the capabilities of embodied AI models in handling complex object arrangement scenarios, with a particular focus on long-horizon robotic manipulation. Extensive experimental results indicate significant room for improvement in existing models in these scenarios. We anticipate that this benchmark will greatly advance research and application development in this field, providing robust support for the advancement of robotic technology in practical applications.

In summary, the contributions of this paper are as follows:
\begin{itemize}
\item We introduce the \textit{RoboCAS} benchmark, specifically designed to train and evaluate robotic manipulation capabilities in complex object arrangement scenarios. To our knowledge, this is the first benchmark focused on such scenarios.
\item This benchmark provides an automated and rapid demonstration generation method, greatly simplifying the data generation process. 
\item Experimental results on baseline models indicate that current embodied AI models still face numerous challenges when handling grasping tasks in complex object arrangement scenarios.
\end{itemize}


\section{Related works}

\subsection{Language-conditioned robot learning}

Recent research interest on embodied AI has focused on accomplishing long-horizon tasks under the condition of human language~\cite{wang2023mimicplay, rt12022arxiv, roboflamingo, dnact, zheng2022vlmbench, huang2023diffusion, bc-z, genh2r, liu2024robouniview}, usually through behavior cloning or reinforcement learning methods to generate robot action policy on the condition of the current environment observation and the language instruction. However most of the models are trained on tasks that is much easier than that in real world, and can hardly be applied to real life or commercial scenes which is much more complex than that in the training datasets on both environment variations and object relations. One line of methods~\cite{rt2, LanguageBind, gr1} try to leverage more easily available video or language data to pre-train the state tokenizers and then fine-tune the whole model with small amount of robot data, to reduce the difficulty of learning robot task, but still perform poorly with occluded target objects because of lacking the relationship between robot action and object reaction. Another cluster of approaches~\cite{long2023discuss, brohan2023can, mu2024robocodex, wu2023tidybot} involve breaking down the long-horizon task into simpler basic tasks using large language models (LLMs). However, these methods still struggle with dense clutter scenes as LLMs are unable to comprehensively understand the relationship between objects in the scene through textual description. To this end, a dataset comprising manipulation tasks in cluttered scenes with densely arranged objects is required to train deployable BC models, which is the starting point of our \textit{RoboCAS}.


\subsection{Datasets and benchmarks for robot manipulation}

A number of datasets for robot learning have been released in the community to facilitate the training of robot skills~\cite{rt-x}, among which grasping~\cite{graspnet1b, eppner2021acronym, zhu2023fanuc, liu2024libero}, pushing~\cite{lynch2023interactive, dasari2019robonet, mees2022calvin} and manipulation~\cite{walke2023bridgedata, fang2023rh20t, xiong2024adaptive} teleoperated by human are most common tasks collected on both simulated and real robot platforms. Unfortunately, in most of the current datasets the scenes are ideally simplified and devoid of variation, while the target objects are considered isolated from other objects while planning and performing the task,
\ie the surrounding obstacles are unrelated to the manipulation policy of the target, which can lead to serious domain gap between scenes and tasks in real world environments when deployed on real robots. Since the datasets collected on real robots~\cite{bharadhwaj2023roboagent, walke2023bridgedata, fang2023rh20t} are inconvenient for researchers to evaluate the trained models, simulation environments~\cite{sapien, makoviychuk2021isaac, todorov2012mujoco} and benchmarks~\cite{mees2022calvin, maniskill, maniskill2, lynch2023interactive} provide a more flexible choice, and our \textit{RoboCAS} follows this line. Different from previous benchmarks, in this paper we focus on the manipulation tasks in cluttered scenes while taking surrounding obstacles around the target object into consideration while generating trajectories, meaning that the policies should tackle the interference among obstacles before directly manipulate the target, which is the most ordinary cases for human-handled tasks in real world. 


\section{The \textit{RoboCAS} benchmark}

To address the challenges robots face when handling complexly arranged objects in environments such as homes or retail spaces, our environment pays more attention on the interference and occlusion of the objects to the actions, as shown in Table~\ref{tab:comparison}, and a specialized dataset is introduced.
This dataset encompasses tasks ranging from simple object manipulation to complex multi-step tasks like searching, each meticulously designed. 
Unlike other benchmarks, this dataset specifically emphasizes spatial reasoning among objects and encourages robots to enhance their ability to predict chain reactions and manipulate under conditions of incomplete observation based on vague language instructions.

\begin{table}[t] 
    \caption{Comparison with existing benchmarks. Lang: language goal annotations. Traj: task execution trajectories. PhyGp: physical grasp simulation. MultiCam: multiple observation cameras. Int: interference of surrounding obstacles. RO: real object models. Scenes: reasonable task executing scenes. Occ: Severe occlusion between objects.}
    \label{tab:comparison}
    \centering
    \begin{tabular}{l|cccccccc}
        \hline 
        Benchmark & Lang & Traj & PhyGp & MultiCam & Int & RO & Scenes & Occ \\
        \hline 
        Language Table~\cite{lynch2023interactive} & \yes & \yes & \no & \no & \yes & \no & \no & \no \\
        ManiSkill~\cite{maniskill, maniskill2} & \no  & \yes  & \yes & \yes & \no & \yes & \no & \no \\
        RLBench~\cite{rlbench} & \yes & \yes & \yes & \yes & \no & \yes & \no & \no \\
        Scaling Up~\cite{ha2023scalingup} & \yes & \yes & \yes & \yes & \no & \yes & \no & \no \\
        Calvin~\cite{mees2022calvin} & \yes & \yes & \yes & \yes & \no & \no & \yes & \no \\
        GraspNet-1Billion~\cite{graspnet1b} & \no & \no & \yes & \yes & \yes & \yes & \no & \yes \\
        ACRONYM~\cite{eppner2021acronym} & \no & \no & \yes & \yes & \yes & \yes & \no & \yes \\
        RoboGen~\cite{wang2023robogen} & \yes & \yes & \no & \yes & \no & \yes & \no & \no \\
        BEHAVIOR~\cite{srivastava2022behavior, li2023behavior} & \no & \yes & \yes & \no & \yes & \yes & \yes & \yes \\
        AI2-THOR~\cite{kolve2017ai2} & \yes & \yes & \no & \yes & \yes & \yes & \yes & \no \\
        \hline
        Ours & \yes & \yes & \yes & \yes & \yes & \yes & \yes & \yes \\
        \hline 
    \end{tabular}

\end{table} 

\subsection{Simulation environment}
\label{sec:env}

\begin{figure}[t]
    \centering
    \includegraphics[width=0.9\linewidth, trim=30 150 30 150, clip]{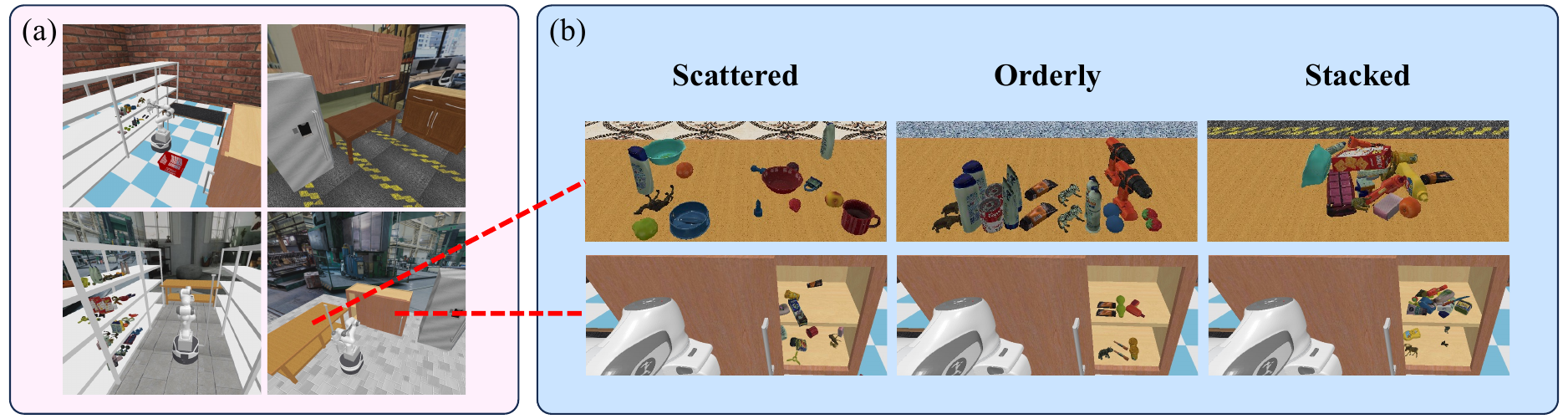}
    \caption{Environment setups of \textit{RoboCAS}. (a) Our environment can provide numerous of scenes by just editing several parameters in configuration files. (b) Three types of scene layouts for manipulable objects are supported: scattered, orderly, and stacked. Due to variations in distances between objects, different layouts present distinct grasping methods and challenges.} 
    \label{fig:envsetup}
\end{figure}

To automatically generate demonstrations, we develop the benchmark within the simulated environment, as illustrated in Figure~\ref{fig:envsetup}.
\textbf{Simulator}: We employ the SAPIEN~\cite{sapien} simulator to construct the task environments, which seamlessly integrates the PhysX physics engine and the Vulkan rendering engine. This combination provides the highly realistic physics simulation coupled with superior rendering quality, enabling robots to accurately learn and perform tasks that closely mimic real-world scenarios.
\textbf{Agent}: The agent used in our task environments is a 7-DoF Franka Emika Panda robot arm, mounted on a mobile base to enhance flexibility. For comprehensive observation, three RGB-D cameras are strategically positioned: one on the robot's head, one on the gripper, and one on the ground. This setup ensures robust observational capabilities, akin to those found on actual robotic platforms.
\textbf{Environmental Objects}: The environmental objects, such as tables, drawers, and backgrounds, primarily originate from the PartNet~\cite{partnet} dataset.We parameterize this setup by specifying the environmental objects and their poses in configuration files. This method not only facilitates the creation and modification of environments but also enables the automatic generation of diverse scenarios. Additional details are provided in Section~\ref{sec:configuration}.
\textbf{Manipulable Objects}: In our simulated environment, we carefully select 46 daily objects, such as toothpaste and cookie box, scanned from real commodities by Fang \etal~\cite{graspnet1b}. These objects, serving as the manipulable objects for the tasks, significantly enhance the authenticity of our environment.

\subsection{Data collection}
Compared to other benchmarks, we design more diverse and rich manipulation scene layouts, and quickly acquire a large number of demonstrations through a cost-effective method using simple task-related scripted policies.

\subsubsection{Scene configuration}
\label{sec:configuration}

The first problem is how to effectively generate various scenes to achieve realistic and reasonable layouts. We manually design a large number of scene templates containing environmental objects, such as cabinets and shelves, and manipulable objects, as shown in Figure~\ref{fig:envsetup}(a). Although these templates are manually designed, the operation process is not complicated, usually just requiring simple edits to the configuration list mentioned in Section~\ref{sec:env}. To further simplify the scene configuration, we initially annotate the operable parts of environmental objects, such as the areas where objects can be placed and the positions of drawer handles, and use the force closure metric (referenced in \cite{graspnet1b}) to determine feasible manipulation poses for manipulable objects. This one-time annotated information can later be automatically retrieved and reused in the scene templates, thereby eliminating the need for manual repetitive coding.

Moreover, our scene templates are divided into three types of scene layouts, used for arranging model objects in various levels of complexity. 
as shown in Figure~\ref{fig:envsetup}(b):
\textbf{a) Scattered scene}: Objects are randomly placed in a single layer on the surface of the operational platform (such as tabletops, inside cabinets), allowing objects to adopt various poses to increase consistency with reality;
\textbf{b) Orderly scene}: Objects are aligned along the surface edge, sorted into columns, simulating the way goods are displayed, in which occlusion and the selection of repeated objects become an issue for the agents;
\textbf{c) Stacked scene}: A large number of unrelated objects are stacked above the target object, creating significant occlusion of observation.
By integrating these different scene layouts, our environment not only increases in diversity and complexity but also better mimics real-world scenarios, providing challenges for effective manipulation.

\subsubsection{Task design}
Corresponding to the aforementioned three scene layouts, we design three types of manipulation tasks that require robots to have long-term planning capabilities, continuously exploring and clearing obstacles during operation.

\begin{figure}[t]
    \centering
    \includegraphics[width=0.9\linewidth, trim=20 215 20 215, clip]{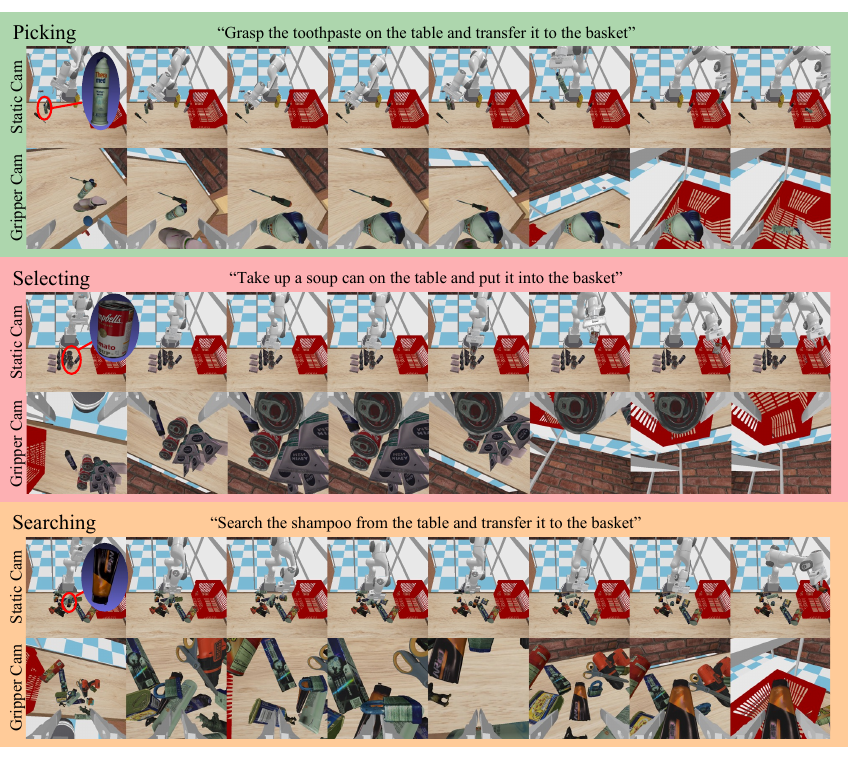}
    \caption{
        The three types of tasks supported in the \textit{RoboCAS} benchmark.
        Picking: Pick up the specified target and move it to the designated location.
        Selecting: Choose and grasp a specific target from multiple identically arranged targets.
        Searching: Find a partially obscured specific target in a stacked scene, clear any obstacles, and then grasp it.
    }
    \label{fig:tasks}
\end{figure}

\paragraph{Picking task.} This task requires the robot to pick up a specified target in the scattered scene and move it to a predetermined destination, providing a foundational operation for our other two tasks. For example, a potential language instruction could be: "Pick up the toothpaste from the table and place it in the red basket." During the grasping action, it is essential to ensure there is sufficient space between objects to avoid contact with surrounding items. The criterion for successful completion of the task is that the robot successfully picks up the target object through physical simulation and accurately places it in the designated container.

\paragraph{Selecting task.} The Selecting task is slightly different from the picking task, requiring the robot to grasp a specific one among multiple identical objects, such as the nearest target, typically found in orderly scenes like goods on shelves. This task primarily tests the robot's ability to understand ambiguous language instructions, for example: "Pick up the nearest toothpaste from the table and place it in the red basket." The success criteria for this task are consistent with those of the picking task.

\paragraph{Searching task.} In this task, the robot's goal is to grasp and relocate a partially obscured target. Before operation, the robot needs to search and locate the target from a partially observed environment while clearing the surrounding area to provide necessary space for subsequent manipulation. Occlusion in the scene brings ambiguity to the task since instructions like "Search out the apple from the heap on the table" cannot directly indicate the exact location of the target. This is a key feature of the task, typically occurring in piled scenes, as shown in Figure~\ref{fig:envsetup}(b). The success criteria for this task are similar to those of the grasping task.


\subsection{Trajectory generation}

\begin{figure}[t]
    \centering
    \includegraphics[width=0.8\linewidth, trim=110 105 110 105, clip]{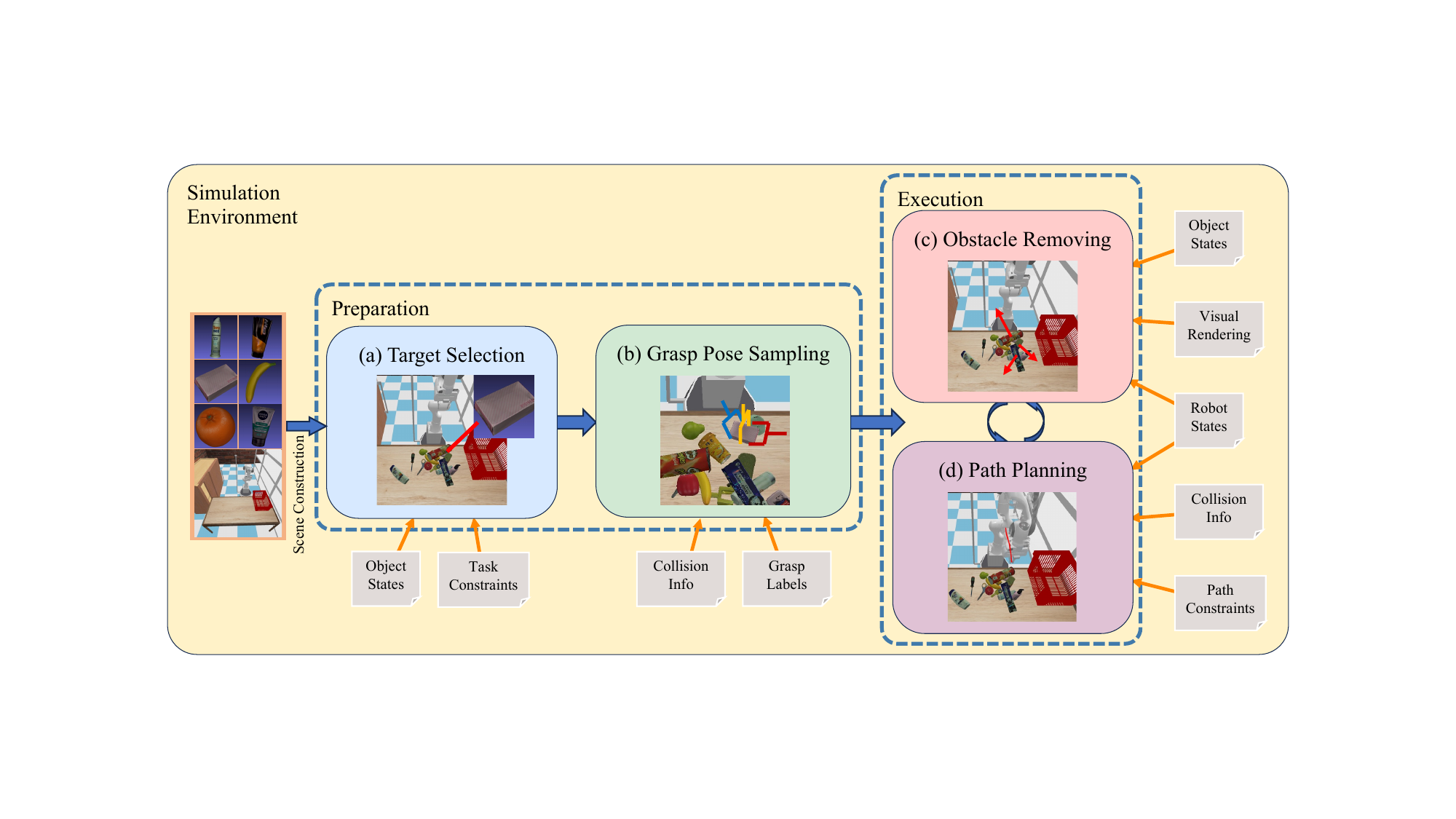}
    \caption{
        The generation process of our \textit{RoboCAS} benchmark:
        (a) Target Selection: After initializing the environment based on the scene template, a target is randomly selected from objects that meet the task's requirements for target state.
        (b) Grasp Pose Sampling: After filtering through collision detection and kinematic calculations of the agent, an appropriate grasp pose is selected from the force-closure annotations.
        (c) Obstacle Removal: Obstacles that may hinder the operation of the target are removed using methods such as pushing or flicking.
        (d) Path Planning: The agent uses the RRT-Connect algorithm to plan a collision-free path to the specified end effector (EEF) pose, integrating collision information provided by the simulator.
    }
    \label{fig:generation}
\end{figure}

After generating scene layouts by modifying scene configuration, we design scripted policies for each task to get demonstration trajectories with access to global information, significantly accelerating data collection compared to manual collection. To simplify the planning process, each type of tasks is divided into several sub-tasks: target selection, grasp pose sampling, obstacle removal, and path planning, as shown in Figure~\ref{fig:generation}. 
Note that in the target selection phase, additional selection criteria can be used such as the distance to the agent among all similar objects and the percent of visible area $p_{vis}$ of each manipulable object:
\begin{equation}
    p_{vis} = S_{vis} / S_{full},
    \label{equ:visibility}
\end{equation}
where $S_{vis}$ is the visible area of the object from a camera $C$, and $S_{full}$ is the full area projected to $C$. Especially, in the searching task one object is selected as the target only if $S_{vis}$ is less than threshold $p_{th}$ when observing from the top-down camera $C_{sel}$ above the workspace, ensuring that the agent should perform the search action before grasping. Language instruction is generated by filling the pre-defined templates after target selection.
In the following grasp sampling phase, grasp poses are sampled from the annotated labels of the target and projected to the current scene, after which collection detection are performed between scene objects and gripper models placed at the these poses to filter out feasible manipulation poses.
In the obstacle removing phase of the searching task, obstacles around the target are detected using the nearest distance between models and visibility score in Equation~\ref{equ:visibility}, 
afterwards a removal plan is generated. The obstacle $O_{obt}$ located at the highest position is prioritized for removal, by horizontally pushing it at point $p_{push}$ on the surface of $O_{obt}$ alongside the direction of vector $v_{push}$:
\begin{equation}
    \begin{aligned}
        v_{push} &= \text{normalize} \left( p_{obt} - p_{tgt} \right) \in \mathbb{R}^{3}, & s.t. \quad v_{push}\left[ z \right]=0, \\
        p_{push} &= p_{obt} - v_{push} * d_{ext}, & s.t. \quad p_{push} \in O_{obt},
    \end{aligned}
\end{equation}
where $p_{tgt}$ is the mass center of the target and $p_{obt}$ is that of $O_{obt}$ and at the height of $p_{tgt}$ (\ie $p_{obt} \left[ z \right] = p_{tgt} \left[ z \right]$), $d_{ext}$ is the extent of $O_{obt}$ from $p_{obt}$ to its verge alongside $-v_{push}$.
The removal plan is executed by the path planning phase by first moving the EEF to $p_{push}$ with finger closed and then pushing $O_{obt}$ along $v_{push}$ for a distance of $2 \times d_{ext}$. The obstacle removing and path planning phases will repeat until the number of detected obstacles is less than 2.
The environment will be reset after several collection episodes, to maintain the scene in a desired state that the planner should consider the interaction between objects while collecting data, which is the main difference between our \textit{RoboCAS} and existing benchmarks. 

\section{Experiments}

\subsection{Experimental setup}

To validate the manipulation capabilities of state-of-the-art (SOTA) models in environments with complex object arrangements, we select two representative models as our baselines: RT-1~\cite{rt12022arxiv}, the most well-known among small models, and RoboFlamingo~\cite{roboflamingo}, which is based on the Multimodal Large Language Models (MLLMs). In both models, we use sequences of RGB images $I^{g}$, $I^{b}$ from the gripper and base cameras, along with language instructions $l$, as inputs to train the manipulation strategies. The models $\Psi_{\theta}$ output a series of 7-dimensional actions $a_{i} \in \mathbb{R}^{7}$ at different time steps based on the given historical observation states $o_{t} = \left( I_{t}^{g}, I_{t}^{b} \right)$:
\begin{equation}
    \left( a_{t}, a_{t-1}, \cdots, a_{t-T} \right) = \Psi_{\theta} \left( o_{t}, o_{t-1}, \cdots, o_{t-T}, l \right),
\end{equation}
where $T$ represents the observation horizon. The actions $a_{i} = \left( p_{i}^{rel}, e_{i}^{rel}, g_{i} \right)$ include $p_{i}^{rel}, e_{i}^{rel} \in \mathbb{R}^{3}$, which represent the relative position and Euler angle shifts of the end effector (EEF) with respect to the current pose, and $g_{i} \in \left\{ 0, 1 \right\}$ is the binary command for opening or closing the gripper fingers.  For implementation details of each model, please consult the respective original publications.

We generated 7300 episodes of trajectories on single PC in 5 days and used for model training, and details can be found in Appendix.
In the RT-1 model, the observation horizon $T$ is set to 5 to provide background for the current trajectory, while the RoboFlamingo model implicitly adopts $T=11$ through an RNN. Multiple steps of action can be generated at each prediction, but only the action corresponding to the current time step is executed. All models are trained using 8 Nvidia A100 GPUs, and they are trained and tested in three different scenarios to ensure that the testing environments correspond to the training environments.

\subsection{Experimental results}

\subsubsection{Overall results}

\begin{table}[h]
    \caption{success rate across different scenes or tasks. (Unit: \%)}
    \label{tab:general-result}
    \centering
    \begin{tabular}{l|ccc}
        \hline
        \multirow{2}{*}{Model}  & Scattered Scene   & Orderly Scene     & Stacked Scene \\
                                & (Picking Task)    & (Selecting Task)  & (Searching Task) \\
        \hline
        RT-1~\cite{rt12022arxiv}& 34.2              & 23.3              & 10.2 \\
        RoboFlamingo~\cite{roboflamingo} & 15.6     & 13.3              & 0 \\
        \hline
    \end{tabular}
\end{table}

As illustrated in Table~\ref{tab:general-result}, we conduct evaluations of the RT-1~\cite{rt12022arxiv} and RoboFlamingo~\cite{roboflamingo} models across a variety of object arrangements. These models exhibit promising success rates in scattered and orderly scenes (34.2\% vs 15.6\%, and 23.3\% vs 13.3\%), demonstrating the efficacy of language-conditioned manipulation models in handling diverse objects within relatively straightforward object arrangement scenarios—situations that are prevalent in real-world environments. However, the success rates in stacked scenes (10.2\% vs 0\%) are notably lower, which not only highlights the baseline models' limitations in spatial reasoning, understanding inter-object interactions, and predicting chain effects but also emphasizes the critical need for introducing such benchmarks to further advance our understanding and capabilities in this area.


\subsubsection{Detailed analysis}

\begin{table}[h]
    \caption{
        Detailed performance on major steps.
        RA: The gripper \textbf{R}eaches \textbf{A}ny manipulable object in the scene.
        PA: \textbf{A}ny manipulable object is \textbf{P}icked up.
        RR: The gripper \textbf{R}eaches the \textbf{R}ight object.
        CO: The agent completes the search action, which includes \textbf{C}learing \textbf{O}bstacles.
        (Unit: \%)
    }
    \label{tab:step-result}
    \centering
    \begin{tabular}{l|ccc|ccc|cccc}
        \hline
        \multirow{2}{*}{Model} & \multicolumn{3}{c|}{Scattered Scene} & \multicolumn{3}{c|}{Orderly Scene} & \multicolumn{4}{c}{Stacked Scene} \\
        & RA & PA & RR & RA & PA & RR & CO & RA & PA & RR \\
        \hline
        RT-1~\cite{rt12022arxiv} & 84.6 & 48.7 & 36.1 & 79.4 & 44.1 & 30.8 & 7.7 & 71.8 & 28.2 & 22.2 \\
        RoboFlamingo~\cite{roboflamingo} & 91.1 & 31.1 & 37.8 & 93.3 & 37.8 & 44.4 & 3.2 & 41.3 & 9.5 & 9.5 \\
        \hline
    \end{tabular}
\end{table}

To deepen our understanding of the results presented in Table~\ref{tab:general-result}, we detail the success rates of key steps across various scenarios or tasks in Table~\ref{tab:step-result}. For example, in scattered settings, the success rate of an RT-1 driven agent reaching any manipulable target from its initial position is 84.6\%, reflecting the agent's ability to orient and progressively approach the target based on language instructions. From Table~\ref{tab:step-result}, it is evident that although agents can reach manipulable objects in most cases, the probability of correctly reaching the target object is relatively low (e.g., 84.6\% vs 48.7\% or 79.4\% vs 44.1\%). This indicates difficulties in model's ability to identify target objects from language instructions, especially in stacked scenarios (71.8\% vs 28.2\%) where the obstruction of target objects adds complexity to aligning instructions with target observations,
as well as the need for the ability to actively explore unobserved scenarios.
Notably, compared to RT-1, which only uses the pre-trained EfficientNetB3~\cite{tan2019efficientnet} as an image tokenizer, RoboFlamingo, which utilizes the pre-trained CLIP~\cite{radford2021learning} model to bind multi-modal features, exhibits the relatively higher language-vision alignment capability (91.1\% vs 84.6\%). This emphasizes the importance of collaborative training across different modalities for foundational models in manipulation tasks. 

\begin{figure}[t]
    \centering
    \includegraphics[width=1.0\linewidth, trim=150 220 150 220, clip]{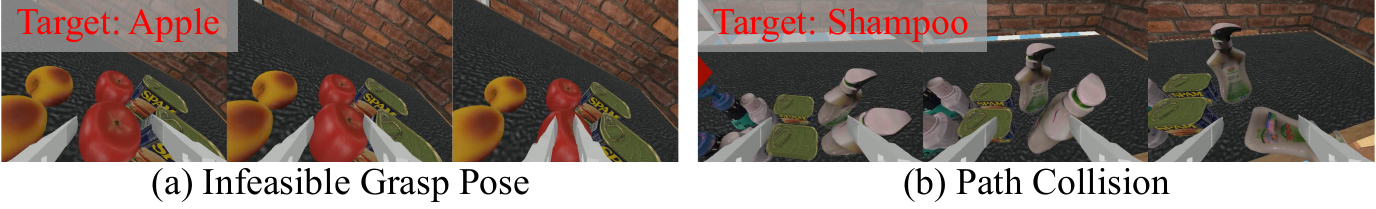}
    \caption{
        Failure cases occurred in our experiments.
        (a) Infeasible grasp pose that lead to the slipping of the target object.
        (b) Failure cause by the collision between target and agent and the corresponding pose change.
    }
    \label{fig:failure}
\end{figure}

In addition to target identification issues, spatial reasoning and predicting chain reactions also present significant challenges in these tasks. For instance, as shown in Figure~\ref{fig:failure}, the agent moves the gripper next to an apple but overlooks the object's width and orientation, resulting in the apple slipping when the fingers close; during the process of grasping shampoo, the ignorance of collision between the target and agent in motion plan causes the displacement of the target and the subsequent failure.
Given the inherent limitations of RGB images in perceiving 3D geometries, utilizing information such as depth and point clouds, along with corresponding models to establish a correlation between geometry, spatial configuration and action, is crucial for enhancing performance in interaction-rich tasks and cluttered scenes. Overall, the experimental results reveal the limitations of current robotic manipulation models, including the ability to reason spatially and predict chain reactions under ambiguous instructions.

\subsection{Limitations and future work}

This paper primarily focuses on the language-based manipulation capabilities of robots in environments with complex object arrangements, but does not address mobile bases and navigation strategies, limiting the variety and flexibility of tasks that robots can perform. This will be a key area of focus for our future work. Additionally, due to the lack of the hardware environment, we have not been able to validate our data in the real world to explore the significant potential benefits it may bring to practical applications (Sim2Real).


\section{Conclusion}

In this paper, we introduce the novel benchmark named \textit{RoboCAS}, which focuses on the active exploration and manipulation of robots in environments with complex object arrangements based on language instructions for long-horizon tasks such as searching. This benchmark employs the streamlined scripted policies to automatically and rapidly collect demonstrations featuring manipulable objects in scattered, orderly, and stacked arrangements, all sourced from real-world scans. Detailed experiments on baseline models reveal the limitations of state-of-the-art models in manipulation tasks within complex arrangement environments. These limitations primarily include inadequate handling of ambiguous instructions, a lack of spatial reasoning capabilities, and insufficient prediction of chain effects. These findings highlight the essential direction for further research on foundational models in the field of embodied AI.


\newpage

\bibliographystyle{unsrt}

\section*{Checklist}


\begin{enumerate}

\item For all authors...
\begin{enumerate}
  \item Do the main claims made in the abstract and introduction accurately reflect the paper's contributions and scope?
    \answerYes{}
  \item Did you describe the limitations of your work?
    \answerYes{}
  \item Did you discuss any potential negative societal impacts of your work?
    \answerNA{}
  \item Have you read the ethics review guidelines and ensured that your paper conforms to them?
    \answerYes{}
\end{enumerate}

\item If you are including theoretical results...
\begin{enumerate}
  \item Did you state the full set of assumptions of all theoretical results?
    \answerNA{}
  \item Did you include complete proofs of all theoretical results?
    \answerNA{}
\end{enumerate}

\item If you ran experiments (e.g. for benchmarks)...
\begin{enumerate}
  \item Did you include the code, data, and instructions needed to reproduce the main experimental results (either in the supplemental material or as a URL)?
    \answerNo{The full data is not publicly accessible due to the constraints imposed by the enterprise. One can generate your own dataset through the provided code.}
  \item Did you specify all the training details (e.g., data splits, hyperparameters, how they were chosen)?
    \answerNo{One can split the dataset generated by yourself.}
  \item Did you report error bars (e.g., with respect to the random seed after running experiments multiple times)?
    \answerNo{}
  \item Did you include the total amount of compute and the type of resources used (e.g., type of GPUs, internal cluster, or cloud provider)?
    \answerYes{}
\end{enumerate}

\item If you are using existing assets (e.g., code, data, models) or curating/releasing new assets...
\begin{enumerate}
  \item If your work uses existing assets, did you cite the creators?
    \answerYes{}
  \item Did you mention the license of the assets?
    \answerYes{License can be found in the original source provided in the code.}
  \item Did you include any new assets either in the supplemental material or as a URL?
    \answerNo{}
  \item Did you discuss whether and how consent was obtained from people whose data you're using/curating?
    \answerYes{Public dataset.}
  \item Did you discuss whether the data you are using/curating contains personally identifiable information or offensive content?
    \answerNA{No such information included.}
\end{enumerate}

\item If you used crowdsourcing or conducted research with human subjects...
\begin{enumerate}
  \item Did you include the full text of instructions given to participants and screenshots, if applicable?
    \answerNA{}
  \item Did you describe any potential participant risks, with links to Institutional Review Board (IRB) approvals, if applicable?
    \answerNA{}
  \item Did you include the estimated hourly wage paid to participants and the total amount spent on participant compensation?
    \answerNA{}
\end{enumerate}

\end{enumerate}

\appendix

\section{Dataset documentation}
\label{sec:documentation}

This dataset contains the generated trajectory of robot manipulation on picking, selecting and searching tasks. You can generate your own dataset through the code provided on \url{https://github.com/notFoundThisPerson/RoboCAS-v0}, and some demonstrations are given on \url{https://huggingface.co/datasets/zlm898/RoboCAS-v0} (DOI: 10.57967/hf/2420). Croissant metadata record: \url{https://huggingface.co/datasets/zlm898/RoboCAS-v0/blob/main/croissant.json}. The directory structure is as follows:
\begin{itemize}
    \item \textbf{data\_info.json} \quad \# Overview of the dataset information, in dictionary format.
    \item \textbf{episode\_0000000} \quad \# Trajectory 0.
    \begin{itemize}
        \item \textbf{episode\_info.npz} \quad \# Actions and robot states in the trajectory.
        \item \textbf{gripper\_camera} \quad \# Data of the camera mounted on the gripper.
        \begin{itemize}
            \item \textbf{cam\_pos\_wrt\_\$\{parent\}.npy} \quad \# Camera pose w.r.t. its parent link.
            \item \textbf{intrinsic.npy} \quad \# Intrinsic of this camera.
            \item \textbf{rgb} \quad \# RGB images in this trajectory.
            \begin{itemize}
                \item \textbf{0000.png}
                \item $\cdots$
            \end{itemize}
            \item \textbf{depth} \quad \# Depth images in this trajectory. Same structure as "rgb".
        \end{itemize}
        \item \textbf{base\_camera} \quad \# Data of the camera mounted on the robot base. Same structure as "gripper\_camera".
        \item \textbf{static\_camera} \quad \# Data of the camera mounted on the ground. Same structure as "gripper\_camera".
    \end{itemize}
    \item \textbf{episode\_0000001} \quad \# Same structure as "episode\_0000000".
    \item $\cdots$
\end{itemize}
In each trajectory folder, the "episode\_info.npz" file contains the trajectory of the agent, the structure and the explanations of each item is as follows:
\begin{itemize}
    \item \textbf{rel\_pos} \quad \# Relative position shift of the EEF w.r.t. the last EEF pose in Cartesian coordinate.
    \item \textbf{rel\_orn} \quad \# Relative orientation shift of the EEF w.r.t. the last EEF pose in Quaternion.
    \item \textbf{ee\_pos} \quad \# Absolute position of the EEF w.r.t. the arm base in Cartesian coordinate.
    \item \textbf{ee\_orn} \quad \# Absolute orientation of the EEF w.r.t. the arm base in Quaternion.
    \item \textbf{robot\_joints} \quad \# Joint angles of the arm.
    \item \textbf{arm\_joint\_vel} \quad \# Velocities of the arm joints.
    \item \textbf{base\_pos} \quad \# Position of the arm base w.r.t. the world in Cartesian coordinate.
    \item \textbf{base\_orn} \quad \# Orientation of the arm base w.r.t. the world in Quaternion.
    \item \textbf{base\_rel\_pos} \quad \# Relative position shift of the arm base w.r.t. the the pose in last step in Cartesian coordinate.
    \item \textbf{base\_rel\_orn} \quad \# Relative orientation shift of the arm base w.r.t. the the pose in last step in Quaternion.
    \item \textbf{gripper\_width} \quad \# Open width of the gripper fingers.
    \item \textbf{gripper\_status} \quad \# Open/close command of the gripper.
    \item \textbf{episode\_length} \quad \# Length of the trajectory.
    \item \textbf{language\_goal} \quad \# Global goal instruction of this trajectory.
    \item \textbf{language\_embedding} \quad \# Embedding of the goal instruction generated by \href{https://huggingface.co/sentence-transformers/all-MiniLM-L6-v2}{Mini LAMMA}.
    \item \textbf{step\_lang\_goals} \quad \# Goal annotation of the sub-task for each step of action in this trajectory.
    \item \textbf{step\_goal\_embs} \quad \# Embedding of step goals generated by \href{https://huggingface.co/sentence-transformers/all-MiniLM-L6-v2}{Mini LAMMA}.
    \item \textbf{step\_goal\_type} \quad \# Type of the sub-task goals in each step.
\end{itemize}
All of the Quaternion in our dataset are stored in "xyzw" order. 
All relative transforms are calculated under the frame of the last time step, \ie $\Delta T_{t} = T_{t-1}^{-1} T_{t}$, where $\Delta T_{t}, T_{t} \in SE(3)$ are the relative action and the pose at time step $t$.
Details can be found in our project. 

\section{Boarder impacts}
Our benchmark and dataset can facilitate the research on long-horizon robot manipulation task under complex scenarios with various-shaped daily objects and cluttered arrangements in a cost-effective manner.

\section{Data sheets for dataset}
\subsection{Motivation}
\begin{enumerate}
    \item For what purpose was the dataset created? \\
    See our Abstract.
    
    \item Who created the dataset (e.g., which team, research group) and on behalf of which entity (e.g., company, institution, organization)? \\
    The authors listed on this paper, which include researchers from Meituan.

    \item Who funded the creation of the dataset? \\
    Meituan.
\end{enumerate}

\subsection{Composition}
\begin{enumerate}
    \item What do the instances that comprise the dataset represent? \\
    Trajectories of robot manipulation on common daily objects (such as fruits, shampoo bottles and toys) performed in simulation environments with complex arrangements. 

    \item How many instances are there in total(of each type,if appropriate)? \\
    This benchmark contains 46 manipulable objects, and the user can generate any number of trajectories with the provided code in our project.

    \item Does the dataset contain all possible instances or is it a sample (not necessarily random) of instances from a larger set? \\
    The provided dataset only contains several demonstrations for each category of task in our paper.

    \item What data does each instance consist of? \\
    Each trajectory is composed of RGB and depth images from different cameras, language instructions and trajectory data, see Appendix~\ref{sec:documentation}.

    \item Is there a label or target associated with each instance? \\
    Yes.

    \item Are there recommended data splits (e.g., training, development/validation, testing)? \\
    User can divide the generated trajectories by himself in certain proportions, \eg 90\% for training and 10\% for validation.

    \item Are there any errors, sources of noise, or redundancies in the dataset? \\
    Each trajectory is sampled randomly at each generation run.

    \item Is the dataset self-contained, or does it link to or otherwise rely on external resources (e.g., websites, tweets, other datasets)? \\
    This dataset is built on assets that are from GraspNet-1Billion and PartNet-Mobility datasets, see our paper for details.
\end{enumerate}

\subsection{Collection process}
\begin{enumerate}
    \item How was the data associated with each instance acquired? \\
    Through the code provided in our project.

    \item What mechanisms or procedures were used to collect the data (e.g., hardware apparatuses or sensors, manual human curation, software programs, software APIs)? \\
    The data is collected in simulation platform SAPIEN.

    \item If the dataset is a sample from a larger set, what was the sampling strategy (e.g., deterministic, probabilistic with specific sampling probabilities)? \\
    The manipulable objects are sampled from the GraspNet-1Billion based on their level of difficulty for graspping using a Panda gripper, as well as the ease of identifying them by name. The models in PartNet-Mobility are sampled based on their reasonability in the environment and the difficulty to be operated. Objects in each scene is sampled randomly.

    \item Who was involved in the data collection process? \\
    Authors of this paper.
\end{enumerate}

\subsection{Preprocessing/cleaning/labeling}
\begin{enumerate}
    \item Was any preprocessing/cleaning/labeling of the data done (e.g., discretization or bucketing, tokenization, part-of-speech tagging, SIFT feature extraction, removal of instances, processing of missing values)? \\
    Yes, see our main paper.

    \item Was the "raw" data saved in addition to the preprocessed/cleaned/labeled data (e.g., to support unanticipated future uses)? \\
    Yes.

    \item Is the software that was used to preprocess/clean/label the data available? \\
    Yes, Python and SAPIEN is available for anyone.
\end{enumerate}

\subsection{Uses}
\begin{enumerate}
    \item Has the dataset been used for any tasks already? \\
    Yes, we use the dataset for training the behavior cloning in manipulation task, as specified in our main paper.

    \item What (other) tasks could the dataset be used for? \\
    The dataset can also be used in video generation besides behavior cloning in our main paper.

    \item Is there anything about the composition of the dataset or the way it was collected and preprocessed/cleaned/labeled that might impact future uses? \\
    No.

    \item Are there tasks for which the dataset should not be used? \\
    No.
\end{enumerate}

\subsection{Distribution}
\begin{enumerate}
    \item How will the dataset will be distributed (e.g., tarball on website, API, GitHub)? Does the dataset have a digital object identifier (DOI)? \\
    The project can be acquired on GitHub as specified in Appendix~\ref{sec:documentation}. DOI: 10.57967/hf/2420.

    \item When will the dataset be distributed? \\
    See repository https://github.com/notFoundThisPerson/RoboCAS-v0

    \item Have any third parties imposed IP-based or other restrictions on the data associated with the instances? \\
    GraspNet-1Billion dataset used in our benchmark is under CC BY-NC-SA 4.0.

    \item Do any export controls or other regulatory restrictions apply to the dataset or to individual instances? \\
    No.
\end{enumerate}

\subsection{Maintenance}
\begin{enumerate}
    \item Who will be supporting/hosting/maintaining the dataset? \\
    The authors of paper will be maintaining the dataset.

    \item How can the owner/curator/manager of the dataset be contacted (e.g., email address)? \\
    Please contact: zhengliming04@meituan.com or yanfeng05@meituan.com.

    \item Is there an erratum? \\
    The data generation policy can be updated in later code versions.

    \item Will older versions of the dataset continue to be supported/hosted/maintained? \\
    Yes, all versions are maintained on GitHub repository.

    \item If others want to extend/augment/build on/contribute to the dataset, is there a mechanism for them to do so? \\
    We provide CC BY-NC-SA 4.0 license for our dataset, so others can freely contribute in our dataset.
\end{enumerate}

\section{License}
The code is under MIT license and the dataset is under CC BY-NC-SA 4.0 license. All rights reserved.

\section{Data generation details}
\label{sec:data-gen-detail}
We generate the trajectory data using SAPIEN simulator on a single PC with a Nvidia RTX 3070 GPU in 5 days.
Assets used in our experiments are from \href{https://graspnet.net/datasets.html}{GraspNet-1Billion} dataset and \href{https://sapien.ucsd.edu/browse}{PartNet-Mobility} dataset. 
In order to emphasizing the limitations of baseline models described in this paper, we use a single table-top environment in training data generation, meanwhile the target information and agent states and actions are recorded.
Totally 2.8k trajectories (260k steps) are generated for picking task, 3k trajectories (275k steps) for selecting task and 1.5k trajectories (285k steps) for searching task.
Ranges of sub-tasks and their goals are also recorded during generation.

\section{Training details}
\subsection{RT-1}
We have implemented the RT-1 model with PyTorch, and trained the models with our generated trajectories under three tasks described in this paper. All models take the two-viewed RGB images from base and gripper cameras as input. The observation horizon used in our experiments is 6, and the corresponding action at each time step is predicted. In all experiments the models are trained using data described in Appendix~\ref{sec:data-gen-detail} with learning rate varying from $10^{-4}$ to $10^{-5}$ with a cosine annealing learning rate schedule. Each model is trained using 8 Nvidia A100 GPUs with batch size 64. Models at the 7th epoch (picking task, out of 10 epochs), the 10th epoch (selecting task, out of 20 epochs) and the 19th epoch (searching task, out of 25 epochs) are used in the experiments. 

\subsection{RoboFlamingo}
The RoboFlamingo model is also trained using data described in Appendix~\ref{sec:data-gen-detail}, and the input data are all the same as that of RT-1. The observation horizon is set to 11. Each model is trained using 8 Nvidia A100 GPUs with batch size 6 and a constant learning rate $10^{-4}$. Models are trained for 10 epochs respectively, and the 6th (picking task), 8th (selecting task) and 10th (searching task) are used in the experiments.

\end{document}